\documentclass{article}

\usepackage{hyperref, color}
\usepackage{amsthm}
\usepackage{epsfig, graphics, graphicx}
\usepackage{latexsym}
\usepackage[parfill]{parskip}
\usepackage[tight]{subfigure}
\usepackage{amsmath,amssymb,enumerate,comment}
\usepackage{accents}
\usepackage{natbib}
\usepackage[margin=1.35in]{geometry}

\newtheorem{theorem}{Theorem}
\newtheorem{corollary}[theorem]{Corollary}

\newcommand{\hi}{\hat{i}}

\let\hat\widehat

\newcommand{\E}{\mathbb{E}}

\newcommand{\lbs}{\tfrac{c}{\sigma_n}}

\title{An Analysis of Active Learning With Uniform Feature Noise}

\author{\\
Aaditya Ramdas$^{12}$\\
\texttt{aramdas@cs.cmu.edu}\\
\hspace{4in} 
\and \\
Barnab\'{a}s P\'{o}czos$^2$\\
\texttt{bapoczos@cs.cmu.edu} \\
\hspace{2in} 
\and \\
Aarti Singh$^2$ \\
\texttt{aarti@cs.cmu.edu} \\
\and \\
Larry Wasserman$^{12}$ \\
\texttt{larry@stat.cmu.edu} \\
\and \\ 
Department of Statistics$^1$ and Machine Learning Department$^2$\\
Carnegie Mellon University\\ 
}

\begin{document}

\maketitle

\begin{abstract}
In active learning,
the user sequentially chooses values for feature $X$
and an oracle returns the corresponding label $Y$.
In this paper, we consider the effect of feature noise
in active learning, which could arise either because $X$ itself is being measured, or it is corrupted in transmission to the oracle, or the oracle returns the label of a noisy version of the query point.
In statistics, feature noise is known as
``errors in variables'' 
and has been studied extensively in non-active settings.
However, the effect of feature noise in active learning has not been studied before. We consider the well-known Berkson errors-in-variables model with additive uniform noise of width $\sigma$.

Our simple but revealing setting is that of one-dimensional binary classification setting where
the goal is to learn a threshold (point where
the probability of a $+$ label crosses half). We deal with regression functions that are antisymmetric in a region of size $\sigma$ around the threshold and also satisfy Tsybakov's margin condition around the threshold. We 
 prove minimax lower and upper bounds which demonstrate that when $\sigma$ is smaller than the minimiax active/passive noiseless error derived in \cite{CN07}, then noise has no effect on the rates and one achieves the same noiseless rates. For larger $\sigma$, the \textit{unflattening} of the regression function on convolution with uniform noise, along with its local antisymmetry around the threshold, together yield a behaviour where noise \textit{appears} to be beneficial.
Our key result is that active learning can buy significant improvement over a passive strategy even in the presence of feature noise.
\end{abstract}

\section{Introduction}

Active learning is a machine learning paradigm where the algorithm
interacts with a label-providing oracle in a feedback driven loop
where past training data (features queried and corresponding labels)
are used to guide the design of subsequent queries. Typically, the
oracle is queried with an exact feature value and the oracle returns
the label corresponding precisely to that feature value. However, in
many scenarios, the feature value being queried can be noisy and it
helps to analyze what would happen in such a setting. Such situations include noisy sensor measurements of features, corrupted transmission of data from source to
storage, or just access to a limited noisy oracle. 

The errors-in-variables model has been well studied in the statistical
literature and their effect can be profound.
In density estimation,
Gaussian error causes the minimax rate
to become logarithmic in sample size instead of polynomial, see \cite{F91}. For results in passive regression, refer to \cite{F93,F09,CRSC10}, and for passive classification, see \cite{LM12}. However, classification has not been studied in the \textit{Berkson} model introduced below. Also, deconvolution estimators require the noise fourier transform to be bounded away from zero, ruling out uniform noise. Finally, to the best of our knowledge, feature noise has not been studied for active learning in any setting.

The {\em classical errors in variables model} has the graphical form $W \leftarrow X \rightarrow Y$, representing
\begin{align*}
W &= X + \delta~,\\
Y &= m(X) + \epsilon~.
\end{align*}
Here, the label $Y$ depends on the feature $X$ but we do not observe $X$;
rather we observe the noisy feature $W$.
The {\em Berkson errors in variables model} is
\begin{align*}
X &= W + \delta~,\\
Y &= m(X) + \epsilon~.
\end{align*}
The difference is that we start with an observed feature $W$ and then noise is added to determine $X$.
Graphically, this model is
$W \rightarrow X \rightarrow Y$.

In this paper,
we focus on the Berkson error model
since it  intuitively makes more sense for active learning - it captures the idea that we request a label for feature
$W$, but the oracle returns the label for $X$ which is a corrupted
version generated from $W$, i.e. the noise occurs between the label request
and the oracle output. We use uniform noise since it yields insightful behavior and also has not been addressed in the literature.
We conjecture that qualitatively similar results
hold for other symmetric error models.

\subsection{Setup}

\paragraph{Threshold Classification.} Let $\mathcal{X}=[-1,1]$, $\mathcal{Y}=\{+,-\}$, and
$f:\mathcal{X}\to\mathcal{Y}$ denote a classification rule.  Assuming
$0/1$ loss, the risk of the classification rule $f$ is
$R(f)=\mathbb{E}[1_{\{f(X)\neq Y\}}]=\mathbb{P}(f(X)\neq Y)$.  It is
known that the Bayes optimal classifier, the best measurable
classifier that minimizes the risk $f^*=\arg\min_{f} R(f)$, has the
following form
\begin{align*}
f^*(x)=\begin{cases}
+& ~\mbox{ if } m(x)\geq 1/2~,\\
-& ~\mbox{ if } m(x)< 1/2~,
\end{cases}
\end{align*}
where $m(x)=\mathbb{P}(Y=+|X=x)$ is the unknown regression function.
In what follows, we will consider the case where the  $f^*$ 
is a threshold classifier, i.e. 
there exists a unique $t \in [-1,1]$ with $m(t) = 1/2$ such that $m(x)<1/2$ if $x<t$, and $m(x)>1/2$ if $x>t$. 

\paragraph{Berkson Error Model.} The model is:
\begin{enumerate}
\item User chooses $W$ and requests label.
\item Oracle receives a noisy $W$ namely $X= W + U$.
\item Oracle returns $Y$ where
$\mathbb{P}(Y=+|X=x) = m(x)$.
\end{enumerate}
We take the noise to be uniform:
$U\sim {\rm Unif}[-\sigma,\sigma]$, where the noise width $\sigma$ is known for simplicity. 

\paragraph{Sampling Strategies.} In \emph{passive sampling}, assume that we are given a batch of $w_i\sim
{\rm Unif}[-1,1]$ and corresponding labels $y_i$ sampled independently of $\{w_j\}_{j\neq i}$ and
$\{y_j\}_{j \neq i}$. In this case, a strategy $S$ is just an estimator $S_n: (W \times Y)^n \rightarrow [-1,1]$ that returns a guess $\hat{t}$ of the threshold $t$ on seeing $\{w_i, y_i\}_{i=1}^n$.

In \emph{active sampling} we are allowed to sequentially
choose $w_i= S_i(w_1,\ldots,w_{i-1},y_1,\ldots,y_{i-1})$,
where $S_i$ is a possibly random function of past queries
and labels, where the randomness is independent of queries and labels. In this case, a strategy $A$ is a sequence of functions $S_i: (W \times Y)^{i-1} \rightarrow [-1,1]$ returning query points and an estimator $S_n: (W \times Y)^n \rightarrow [-1,1]$ that returns a guess $\hat{t}$ at the end.

Let $\mathcal{S}_n^P, \mathcal{S}_n^A$ be the set of all passive or active strategies (and estimators) with a total budget of $n$ labels.

To avoid the issue of noise resulting in a point outside the domain, we make a (Q)uerying assumption:
\begin{enumerate}
\item[(Q).] Querying within $\sigma$ of the boundary is disallowed.
\end{enumerate}

\paragraph{Loss Measure.} Let $\hat{t}=\hat{t}(W_1^n,Y_1^n)$ denote an estimator of $t$ using
$n$ samples from a passive or active strategy. Our task will be to estimate the location
of $t$, where we measure accuracy of an estimator $\hat{t}$ by
a loss function which is the point error $|\hat{t} - t|$.

\paragraph{Function Class.} In the analysis of rates for classification (among others), it
is common to use the {\it Tsybakov Noise/Margin Condition} (see \cite{T04}), to characterize the behavior of $m(x)$ around the threshold $t$. Given constants $c,C$ with $C\geq c$, $k \geq 1$, and noise level $\sigma$, let $\mathcal{P}(c,C,k,\sigma)$ be the set of regression functions $m(x)$ that satisfy the following conditions (T,M,B) for some threshold $t$:
\begin{enumerate}
\item[(T).] $|x-t|^{k-1} \geq |m(x) -1/2| \geq c |x - t|^{k-1}$ 
whenever  $|m(x) - 1/2| \leq \epsilon_0~$ for some constant $\epsilon_0$ 
\item[(M).] $m(t+\delta) - 1/2 = 1/2 - m(t-\delta)$ for all $\delta \leq \sigma$.
\item[(B).] $t$ is at least $\sigma$ away from the boundary.
\end{enumerate}
On adding noise $U$, the point where $m \star U$ ($\star$ means convolution) crosses half may differ from $t$, the point where $m$ crosses half. However, the antisymmetry assumption (M) and boundary assumption (B) together imply that the two thresholds are the same. Getting rid of (M,B) seems substantially  difficult.

When $\sigma=0$, (Q), (M) and (B) are vacuously satisfied, and this is exactly the class of functions and strategies considered in \cite{CN07}. Smaller $k$ means that the regression function is steeper,
which makes it easier to estimate the threshold
and classify future labels (cf. \cite{SS04}). $k=1$ captures a discontinuous $m(x)$ jumping at $t$.


\paragraph{Minimax Risk.} We are interested in the minimax risk under the point error loss :
\begin{equation}
\mathcal{R}_n(\mathcal{P}(c,C,k,\sigma)) = \inf_{S \in \mathcal{S}_n} \sup_{P \in \mathcal{P}(c,C,k,\sigma)} \mathbb{E}|\hat{t} - t|
\end{equation}
where $\mathcal{S}_n$ is the set of strategies accessing $n$ samples. For brevity, $\mathcal{R}_n^P(k,\sigma)$ or $\mathcal{R}_n^A(k,\sigma)$ denotes risk for (P)assive/(A)ctive sampling stratgies $\mathcal{S}_n^P,\mathcal{S}_n^A$.

\paragraph{Notation $\prec,\succ,\asymp,\preceq,\succeq$.} We analyse minimax point error rates in different
regimes of $\sigma$ as a function of $n$ (or equivalently, for a given point error, we can analyse how the sample size $n$ depends on $\sigma$) 
 and we write $\sigma_n$ for emphasis. In this paper, $f_n \prec g_n$ means $f_n/g_n \rightarrow 0$, $f_n \asymp g_n$ means $c_1g_n \leq f_n \leq c_2g_n$ where $c_1,c_2$ are constants, $f_n \preceq g_n$ means $f_n \prec g_n$ or $f_n \asymp g_n$, $f_n \succeq g_n$ means $g_n \preceq f_n$ and $f_n \succ g_n$ means $g_n \prec f_n$.

\section{Main Result and Comparisions}
The main result of this paper is as follows.\\

\begin{theorem}
Under the Berkson error model, when given $n$ labels sampled actively or passively with assumption (Q), and when the true underlying regression function lies in $\mathcal{P}(c,C,k,\sigma_n)$ for known $k,\sigma_n$, the minimax risk under the point error loss is:
\begin{enumerate}
\item $\mathcal{R}^P_n(\mathcal{P}(k,\sigma)) ~\asymp~ \begin{cases}n^{-\frac1{2k-1}} \mbox{\ \ \ \ \ ~ if $\sigma_n \prec n^{-\frac1{2k-1}}$} \\ \sigma_n^{-({k-\frac{3}{2}})}\sqrt{\frac{1}{n}} \mbox{\ otherwise } \end{cases}$
\item $\mathcal{R}^A_n(\mathcal{P}(k,\sigma)) ~\asymp~ \begin{cases}n^{-\frac1{2k-2}} \mbox{\ \ \ \ \ ~ if $\sigma_n \prec n^{-\frac1{2k-2}}$} \\ \sigma_n^{-({k-2})}\sqrt{\frac1{n}} \mbox{\ otherwise } \end{cases}$
\end{enumerate}
\end{theorem}

When $k=1$, $m(x)$ jumps at the threshold, and we interpret the quantity $n^{-\frac1{2k-2}}$ as being exponentially small, i.e. being smaller than $n^{-p}$ for any $p$. We also suppress logarithmic factors in  $n,\sigma_n$. If the domain was $[-R,R]$, the corresponding passive rates are obtained by substituting $n$ by $n/R$, but active rates remain the same upto logarithmic factors in $R$. 

\paragraph{Remark.}
In this paper, we focus on learning the threshold $t$.
This is relevant because the threshold maybe of intrinsic interest, and also of interest for prediction if, for example, future queries could be made with
a different noise model or
can be obtained (with some cost) noise-free. Similar results can be derived for 0/1-risk.

\paragraph{Zero Noise.} When $\sigma=0$, the assumptions (Q,B,M) are vacuously true, and our  class $\mathcal{P}(c,C,k,0)$ matches the class $\mathcal{P}(c,C,k)$ considered in \cite{CN07}, and our rates for $\sigma=0$ i.e. $n^{-\frac1{2k-1}}$ and $n^{-\frac1{2k-2}}$ are precisely the passive and active minimax point error rates in \cite{CN07}. 

\paragraph{Small Noise.} When the noise is small, we get what we expect - the risk does not change with noise as long as the noise itself is smaller than the noiseless error. In other words, as long as the noise is smaller than the noiseless error rate of $n^{-\frac1{2k-1}}$ for passive learning, passive learners will not really be able to notice this tiny noise, and the minimax rate remains $n^{-\frac1{2k-1}}$. Similarly, as long as the noise is smaller than the noiseless error rate of $n^{-\frac1{2k-2}}$ for active learning, active learners will not really be able to notice this tiny noise, and the minimax rate remains $n^{-\frac1{2k-1}}$. Also, the passive rates vary smoothly - at the point when $\sigma_n \asymp n^{-\frac1{2k-1}}$, the rates for small and large noise coincide. Similarly, at the point when $\sigma_n \asymp n^{-\frac1{2k-2}}$, the aforementioned active rates for small and large noise coincide.

\paragraph{Large Noise and Assumption (M).} When the noise is large, we see a curious behaviour of the rates. When $k>2$, the error rates seem to get smaller/better with larger noise for both active and passive learning, and furthermore the noisy rates can also be better than the noiseless rate! This might seem to violate both the information processing inequality, and our intuition that more noise shouldn't help estimation. Moreover,  a noiseless active learner may be able to simulate a noisy situation by adding noise and querying at the resulting point, and get better rates, violating lower bounds in \cite{CN07}.

However, we make the following crucial but subtle observation. Our claimed rates are \textit{not} about a fixed function class - due to assumption (M), the function class changes with $\sigma$, and in fact (M) requires the antisymmetry of the regression function to hold over a larger region for larger $\sigma$. This set of functions is actually getting smaller with larger $\sigma$. Even though the functions can behave quite arbitrarily outside $(t-\sigma,t+\sigma)$, this assumption (M) on a small region of size $2\sigma$ actually helps us significantly.

Given that there is no contradiction to the results of \cite{CN07} or more fundamental information theoretic ideas, there is also an intuitive explanation of why assumption (M) helps when we have large noise. As we will see in a later figure, convolution with noise seems to ``stretch/unflatten'' the function around the threshold. Specifically, for larger $k > 2$, the regression function can be quite flat around the threshold - convolution with noise makes it less flat and more linear - in fact it behaves linearly over a large region of width nearly $2\sigma$. This is true regardless of whether assumption (M) holds - however if (M) does not hold, then the convolved threshold, which is the point where the convolved function crosses half, need not be the original threshold $t$. While dropping assumption (M) will not hurt if we only want to find the convolved threshold, but given that our aim is to estimate $t$, the problem of figuring out how much the threshold shifted can be quite non-trivial.

Hence, large noise ensures a behaviour that is less flat and more linear around the threshold, and assumption (M) ensures that the threshold doesn't shift from $t$. Intuitively this is why (M) and large noise help, and technically there is no contradiction becasue the function class is getting progressively simpler because of more controlled growth around the threshold.

The main takeaway is that in all settings, active learning yields a gain over passive sampling. We now describe the upper and lower bounds that lead to Theorem 1. The case $k=1$ is handled in detail for intuitionb but proofs for $k>1$  are in the Appendix.

\subsection{Simulation of Noise Convolution}

\begin{figure}[h!] 
\centering
\vspace{-0.12in}
\includegraphics[scale = 0.44]{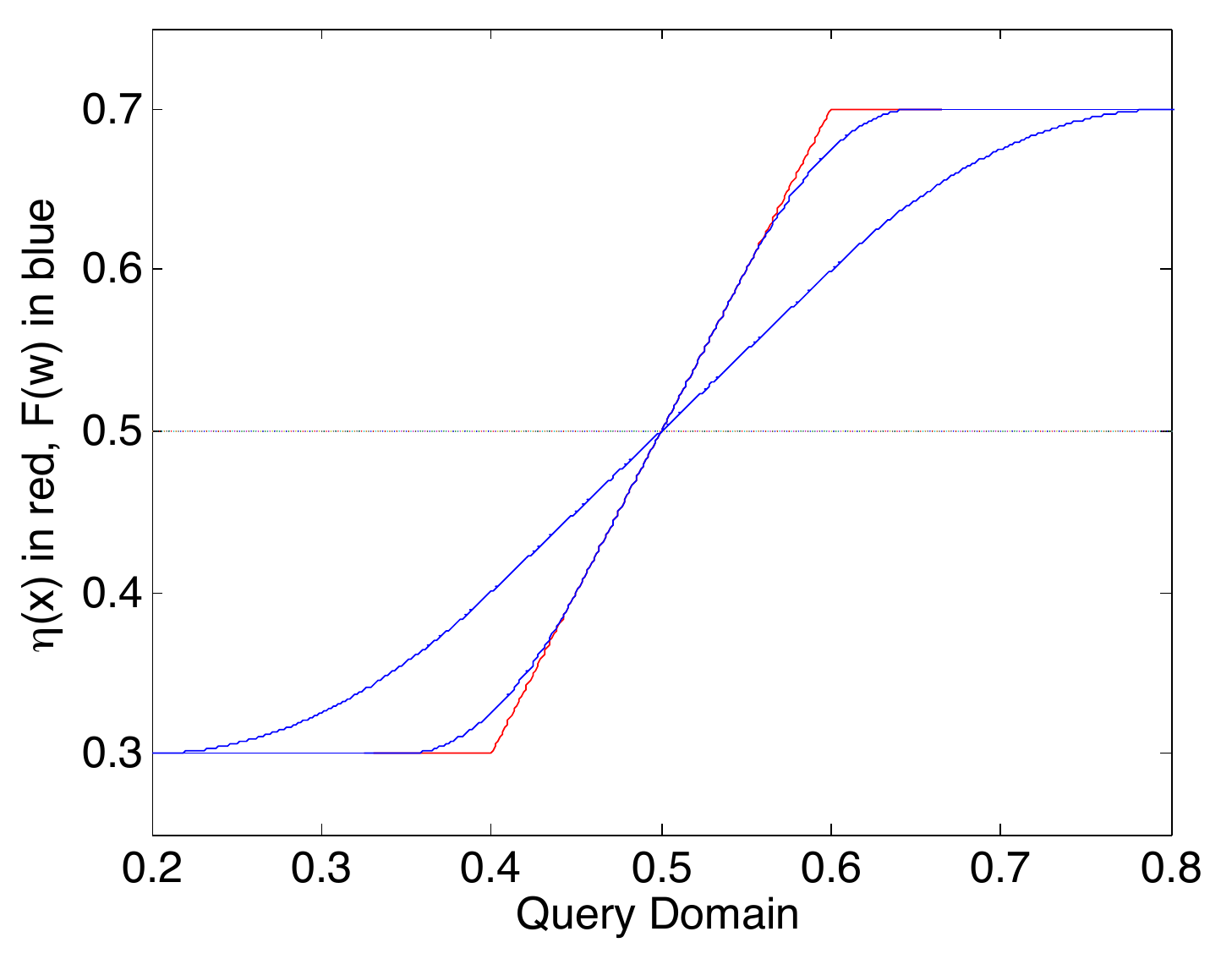}
\includegraphics[scale = 0.44]{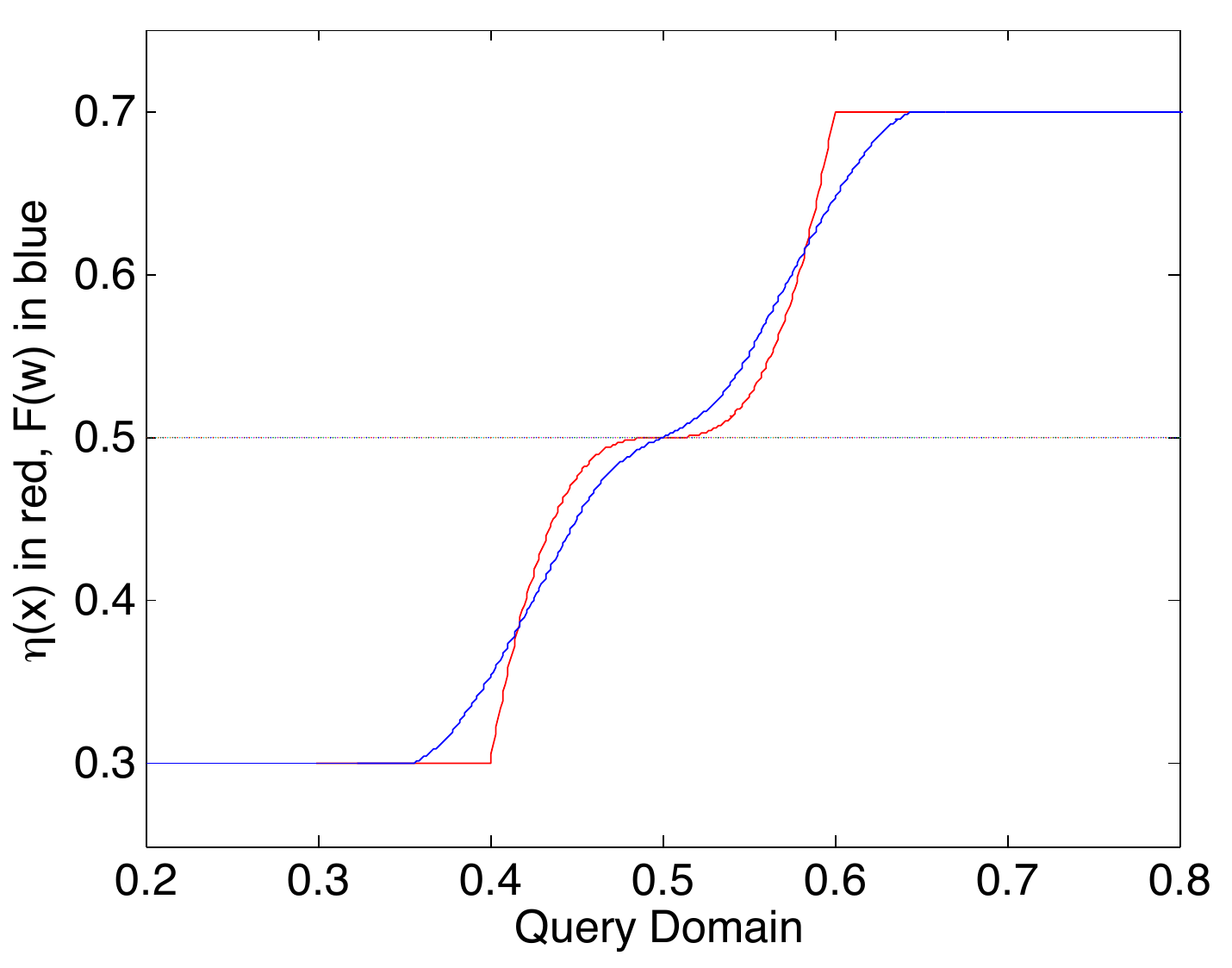}
\includegraphics[scale = 0.44]{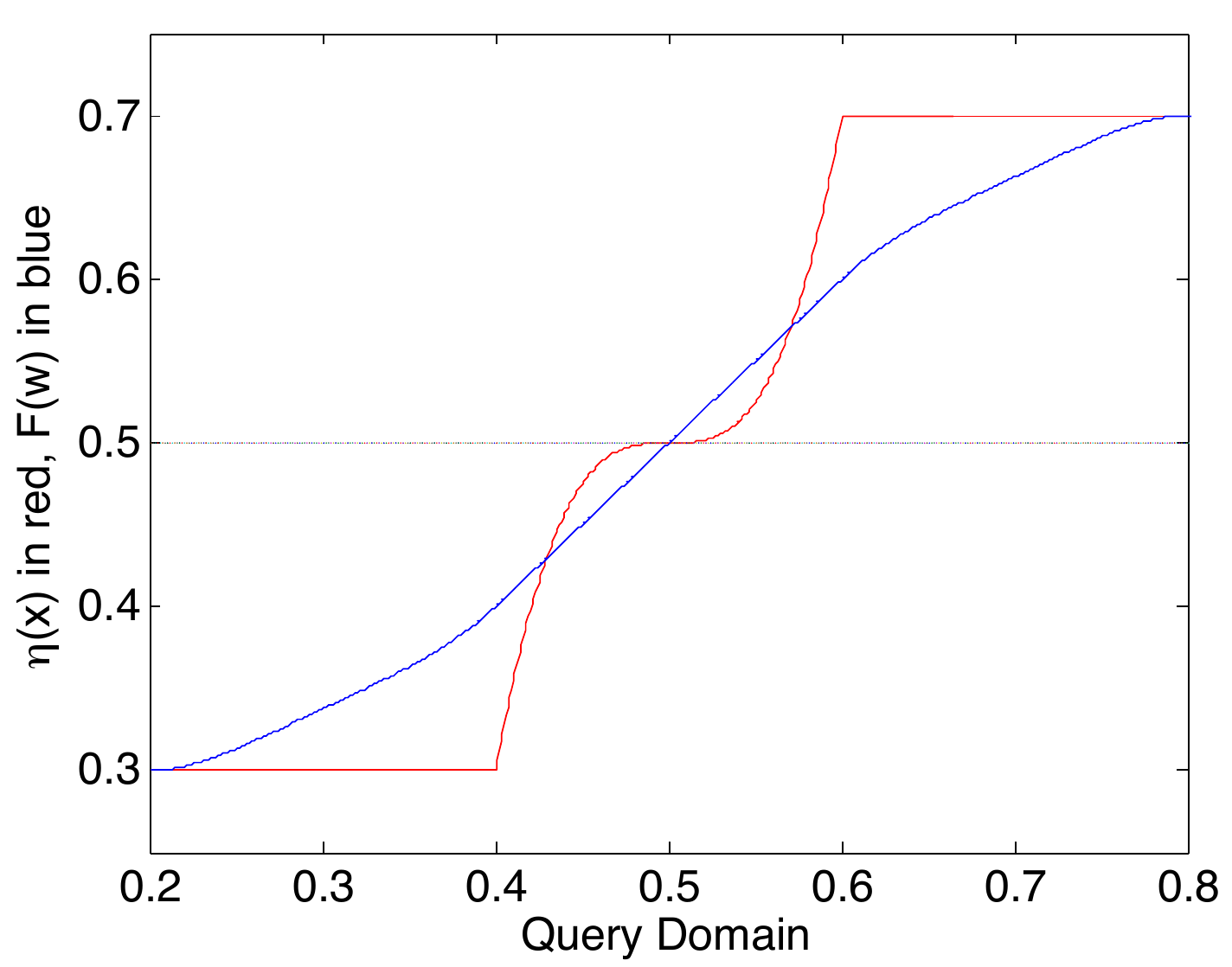}
\vspace{-0.12in}
\caption{Regression function $\eta(x)$ (red) before  and $F(w)$ (blue) after convolution with noise. In all 3 figures, Tsybakov's margin condition holds for $x \in [0.4,0.6]$. The top plot has a linear regression function ($k=2$), and its two blue curves are for $\sigma_n=0.05\ (narrow),0.2\ (wide)$, and they show that a linear growth around $t=0.5$ remains linear. The middle and bottom figure are for a flatter regression function with $k=4$, and $\sigma_n=0.05, 0.2$ respectively, plotted separately for clarity. $k=4$ is harder than for $k=2$ because the red curve is flatter around $t$, making it harder to pinpoint the threshold. However, as one can see in both plots, noise actually \textit{helps} by smoothing it out and making it more linear. However, note that the effect of assumption (M) cannot be understated, due to which in all plots the threshold before and after noise cross half at the same point. 
The effect of noise 
 when $k=1$ can be seen in the following section. 
\label{figk1}
}
\end{figure}

\subsection{Paper Roadmap}

We devote the next two sections to proving the lower and upper bounds, in that order, that lead to Theorem 1. While the proofs will be self-contained, we leave some detailed calculations to the appendix.

For easier readibility, we present lower bounds for $k=1$ first to absorb the technique and then the lower bounds for $k>1$. In Section 2 we will prove \\
\begin{theorem}[Lower Bounds] Under the Berkson error model and assumption (Q),
\begin{enumerate}
\item For $k=1$, the passive/active lower bounds are
\[
\inf_{S \in \mathcal{S}_n^P} \sup_{P \in \mathcal{P}(1,\sigma_n)} \mathbb{E}|\hat{t} - t| ~\succeq~
 \begin{cases}
\frac1{n} \mbox{\ \ \ \ ~ if $\sigma_n \prec \frac1{n}$}\\  \sqrt{\frac{\sigma_n}{n}} \mbox{ \ otherwise}
\end{cases}
\]
\[
\inf_{S \in \mathcal{S}_n^A} \sup_{P \in \mathcal{P}(1,\sigma_n)} \mathbb{E}|\hat{t} - t| ~\succeq~
 \begin{cases}
e^{-n} \mbox{\ ~ if $\sigma_n \prec e^{-n}$}\\  \frac{\sigma_n}{\sqrt{n}} \mbox{ \ otherwise}
\end{cases}
\]
\item For $k>1$, the passive/active lower bounds are
\[
\inf_{S \in \mathcal{S}_n^P} \sup_{P \in \mathcal{P}(k,\sigma_n)} \mathbb{E}|\hat{t} - t| ~\succeq~
 \begin{cases}n^{-\frac1{2k-1}} \mbox{\ if $\sigma_n \prec n^{-\frac1{2k-1}}$} \\ \sigma_n^{-({k-\frac{3}{2}})}\sqrt{\frac{1}{n}} \mbox{\ otherwise } \end{cases}
\]
\[
\inf_{S \in \mathcal{S}_n^A} \sup_{P \in \mathcal{P}(k,\sigma_n)} \mathbb{E}|\hat{t} - t| ~\succeq~
\begin{cases}n^{-\frac1{2k-2}} \mbox{\ if $\sigma_n \prec n^{-\frac1{2k-2}}$} \\ \sigma_n^{-({k-2})}\sqrt{\frac1{n}} \mbox{\ otherwise } \end{cases}
\]\\
\end{enumerate}
\end{theorem}

Following that, we again present active and passive algorithms for $k=1$ first to gather intuition and then generalize them for $k>1$. In Section 3 we will prove \\

\begin{theorem}[Upper Bounds] Under the Berkson error model and assumption (Q),
\begin{enumerate}
\item For $k=1$, a passive algorithm (WIDEHIST) and an active algorithm (ACTPASS) return $\hat{t}$ s.t.
\[
 \sup_{P \in \mathcal{P}(1,\sigma_n)} \mathbb{E}|\hat{t} - t| ~\preceq~
 \begin{cases}
\frac1{n} \mbox{\ \ \ \ ~ if $\sigma_n \prec \frac1{n}$}\\  \sqrt{\frac{\sigma_n}{n}} \mbox{ \ otherwise}
\end{cases}
\]
\[
\sup_{P \in \mathcal{P}(1,\sigma_n)} \mathbb{E}|\hat{t} - t| ~\preceq~
 \begin{cases}
e^{-n} \mbox{\ ~ if $\sigma_n \prec e^{-n}$}\\  \frac{\sigma_n}{\sqrt{n}} \mbox{ \ otherwise}
\end{cases}
\]
\item For $k>1$, a passive  algorithm (WIDEHIST) and an active  algorithm (ACTPASS) return $\hat{t}$ s.t.
\[
 \sup_{P \in \mathcal{P}(k,\sigma_n)} \mathbb{E}|\hat{t} - t| ~\preceq~
 \begin{cases}n^{-\frac1{2k-1}} \mbox{\ if $\sigma_n \prec n^{-\frac1{2k-1}}$} \\ \sigma_n^{-({k-\frac{3}{2}})}\sqrt{\frac{1}{n}} \mbox{\ otherwise } \end{cases}
\]
\[
\sup_{P \in \mathcal{P}(k,\sigma_n)} \mathbb{E}|\hat{t} - t| ~\preceq~
\begin{cases}n^{-\frac1{2k-2}} \mbox{\ if $\sigma_n \prec n^{-\frac1{2k-2}}$} \\ \sigma_n^{-({k-2})}\sqrt{\frac1{n}} \mbox{\ otherwise } \end{cases}
\]\\
\end{enumerate}
\end{theorem}


\section{Lower Bounds}

To derive lower bounds, we will follow the approach of
\cite{IH81,T09} which were exemplified in lower bounds for active
learning problems without feature noise in \cite{CN07,CN08}. The standard methodology is to reduce the problem of classification in the class $P(c,C,k,\sigma)$ to one of hypothesis testing. Similar to \cite{CN07,CN08}, it will suffice to consider two hypotheses and use the following version of Fano's lemma from \cite{T09} (Theorem 2.2).\\

\begin{theorem}[\cite{T09}]\label{fano}
Let $\mathcal{F}$ be a class of models. Associated with each $f \in \mathcal{F}$ we have a probability measure $P_f$ defined on a common probability space. Let $d(.,.): \mathcal{F},\mathcal{F} \rightarrow \mathbb{R}$ be a semi-distance. Let $f_0,f_1 \in \mathcal{F}$ be such that $d(f_0,f_1) \geq 2a$, with $a>0$. Also assume that $KL(P_{f_0},P_{f_1}) \leq \gamma$, where KL denotes the Kullback-Leibler divergence. Then, the following bound holds:
\begin{eqnarray*}
\inf_{\hat{f}}\sup_{f\in\mathcal{F}} P_{f}(d(\hat{f},f)\geq a)
&\geq& \inf_{\hat{t}}\max_{j\in\{0,1\}} P_{f_j}(d(\hat{f},f_j)\geq a)\\
&\geq& \max\left(\frac{e^{-\gamma}}{4},\frac{1-\sqrt{\tfrac{\gamma}{2}}}{2}\right) =: \rho
\end{eqnarray*}
where the $\inf$ is taken with respect to the collection of all possible estimators of $f$ based on a sample from $P_f$.\\
\end{theorem}

\begin{corollary}\label{corrfano}
If $\gamma$ is a constant, then $\rho$ is a constant, and by Markov's inequality, we would get
\begin{equation*}
\inf_{\hat f} \sup_{f \in \mathcal{F}} \mathbb{E}d(\hat f, f) \geq \rho a
\end{equation*}
and the minimax risk under loss $d$ would be $\succeq a$.
\end{corollary}

\paragraph{Proof of Theorem 2, $k=1$.} Choose $\mathcal{F} = \mathcal{P}(1,\sigma_n)$. Let $P_t \in \mathcal{P}(1,\sigma_n)$ denote a regression function with threshold at $t$. 
We choose the semi-metric to be the distance between thresholds, i.e. $d(P_{r}, P_{s}) = |r-s|$. We now choose two such distributions with thresholds at least $2a_n$ apart (we use $a_n$ to explicitly remind the reader that $a$ will later be set to depend on $n$) - let them be denoted $P_{t_0}$ and $P_{t_1}$ with $t_0 = -a_n, t_1 =  a_n$ and
\begin{align*}
P_t(Y=+|X=x) =\begin{cases}
0.5 - c & x < t~,\\
0.5 + c & x \geq t~.
\end{cases}
\end{align*}
Due to addition of noise, we get convolved distributions $P^0 = P_{t_0}(Y|W)$ and $P^1:=P_{t_1}(Y|W)$.

As hinted by the above corollary, we will choose $a_n$ so that $KL(P^0,P^1)$ is bounded by a constant, to get a lower bound on risk $\succeq a_n$. This follows by the following argument from \cite{CN08}.

The $KL(P^0,P^1)$ can be bounded as
\begin{eqnarray}
&& \mathbb{E}^1_{W,Y} \left [ \log \frac{P^1 (W_1^n, Y_1^n)} {P^0 (W_1^n, Y_1^n) } \right ]  \\
&=& \mathbb{E}_{W,Y}^1 
\left [ \log \frac{\prod_i P^1 (Y_i | W_i) P (W_i | W_1^{i-1},Y_1^{i-1}) } 
{ \prod_i P^0 (Y_i | W_i) P (W_i | W_1^{i-1},Y_1^{i-1}) } \right ]  \nonumber \\
&=& \mathbb{E}^1_{W,Y} \left [ \log \frac{\prod_i P^1 (Y_i | W_i) } 
{ \prod_i P^0 (Y_i | W_i) } \right ]  \label{indep}\\
&=& \sum_i \E^1_{W} \left [ \E^1_Y 
\left [  \log \frac{ P^1 (Y_i | W_i) } { P^0 (Y_i | W_i) } \  
\Big | \ W_1,...,W_n  \right ] \right ] \label{passive}\\
&\leq& n \max_{w \in [-1,1]} \E^1_{Y} 
\left [ \log \frac{ P^1 (Y | W) } { P^0 (Y | W) } \ \Big| \ W = w \right ]~ \label{active}\\
&\preceq& n \max_{w \in [-1,1]} (P^1(Y|w) - P^0(Y|w))^2 \label{KLbinom}
\end{eqnarray}
where (\ref{indep}) holds for active learning because the algorithm
determines $W_i$ when given $\{W_1^{i-1},Y_1^{i-1}\}$ and is
independent of the model, and follows by the independence of
future from past for passive learning. (\ref{passive}) holds by law
of iterated expectation.
(\ref{active}) is used for active learning but is not needed for passive learning. (\ref{KLbinom}) follows by  an approximation 
\[
KL(Ber(1/2+p),Ber(1/2+q)) \preceq (p-q)^2
\]
for sufficiently small constants $p,q$.

\begin{figure}[h] 
\includegraphics[scale = 0.3]{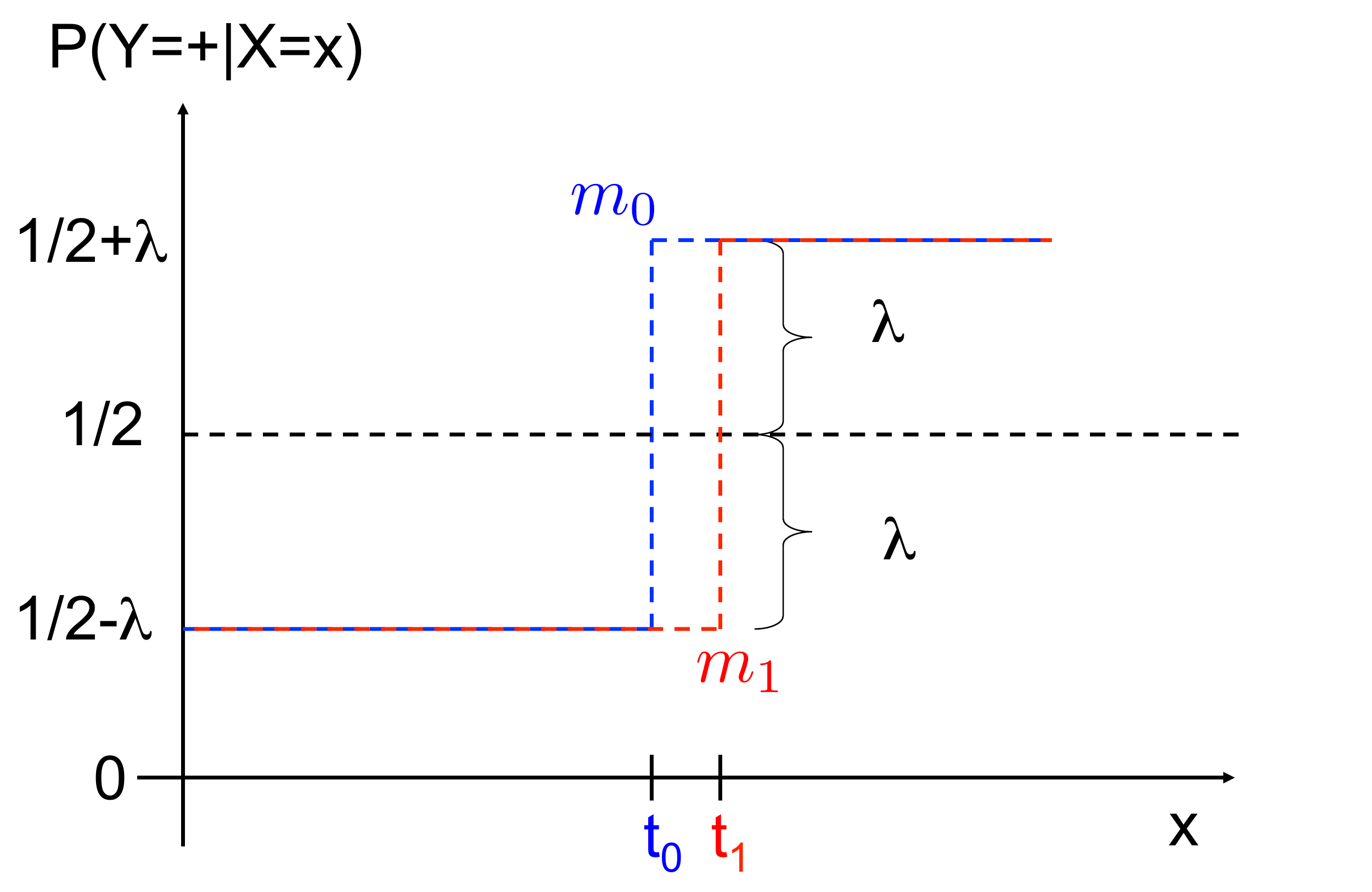}
\includegraphics[scale = 0.3]{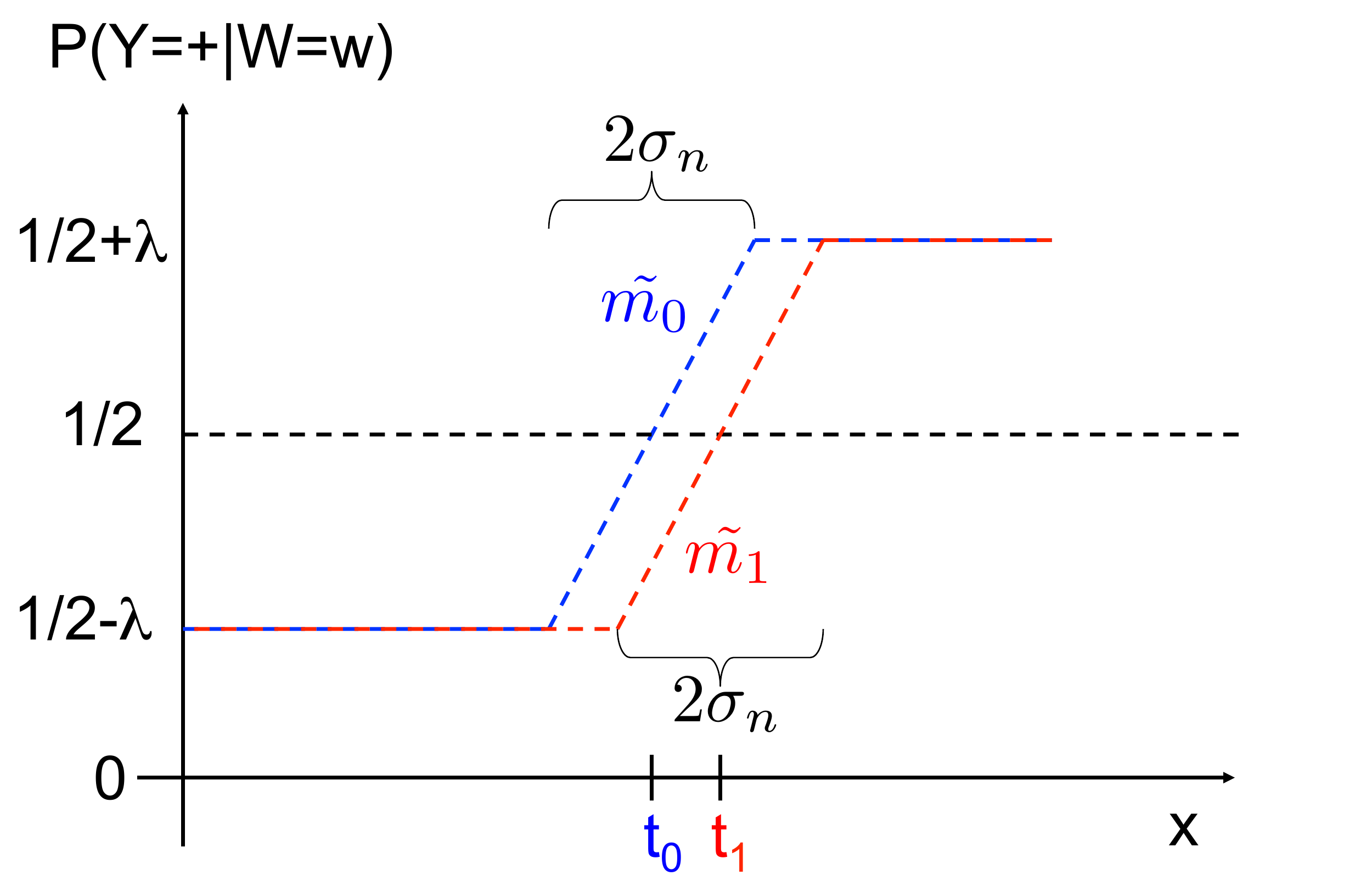}
\caption{Regression functions before (top) and after (bottom) convolution with noise.\label{figlower}
}
\end{figure}

$F_t(w) := P_t(Y|W=w) = \int P_t(Y|X) P(X|W=w) dX$ and a straightforward calculation reveals that 
\begin{align}
\hspace{-0.2in}F_t(w) =  \begin{cases}
0.5 - c  &\mbox{ $w \leq t-\sigma_n$~,} \\
0.5 + \tfrac{c}{\sigma_n}(w-t) &\mbox{$w \in [t-\sigma_n,t+\sigma_n]$~,}\\
0.5 + c  &\mbox{ $w \geq t+\sigma_n$~.}
\end{cases} \label{tildem}
\end{align}

As depicted in Fig.\ref{figlower}, note the behavior before and after convolution with noise:
(i) $m(t) = F(t)=1/2$, hence $F_1(a_n) = 1/2 = F_0(-a_n)$ 
(ii) Both convolved regression functions grow linearly for a 
region of width $2\sigma_n$, and differ only on a width of $2(\sigma_n + a_n)$;
(iii) For a large region $[a_n - \sigma_n, -a_n + \sigma_n]$ of 
size $2(\sigma_n - a_n)$, we have $\big| F_1(w) - F_0(w) \big| = 2a_n c / \sigma_n$, a constant. Their gap varies when $\sigma_n \succeq a_n$ as $\big| F_0(w) - F_1(w) \big|= $ 
\begin{eqnarray*}\label{mdiff}
&& \left\{ \begin{array}{ll}
\Big( w + a_n + \sigma_n \Big)\frac{c}{\sigma_n} &\mbox{$w\in[-a_n-\sigma_n, a_n-\sigma_n]$}\\
2a_n\frac{c}{\sigma_n} &\mbox{$w\in[a_n-\sigma_n,-a_n+\sigma_n]$}\\
\Big( (a_n + \sigma_n) - w \Big)\frac{c}{\sigma_n} &\mbox{$w\in[-a_n+\sigma_n,a_n+\sigma_n]$} \\
0 &\mbox{ otherwise.}
\end{array} \right. \ \ \ 
\end{eqnarray*}
When $\sigma_n \prec a_n$,  $\big| F_1(w) - F_0(w) \big|=$
\begin{eqnarray*}\label{mdiff}
&& \left\{ \begin{array}{ll}
\Big( w + a_n + \sigma_n \Big)\frac{c}{\sigma_n} &\mbox{$w\in[-a_n-\sigma_n, -a_n+\sigma_n]$}\\
2c &\mbox{$w\in[-a_n+\sigma_n,a_n-\sigma_n]$}\\
\Big( (a_n + \sigma_n) - w \Big)\frac{c}{\sigma_n} &\mbox{$w\in[a_n-\sigma_n,a_n+\sigma_n]$} \\
0 &\mbox{ otherwise.}
\end{array} \right. \ \ \ 
\end{eqnarray*}

For active learning, when $\sigma_n \succeq a_n$ we note
$$\max_{w\in[-1,1]} |P^1(Y|w) - P^0(Y|w)| = \frac{2a_nc}{\sigma_n}$$
and get
$
KL(P^0,P^1) \preceq n\frac{a_n^2}{\sigma_n^2}
$
by Eq.(\ref{KLbinom}).
We choose $a_n \asymp \frac{\sigma_n}{\sqrt n}$, which becomes our active minimax error rate by Corollary \ref{corrfano} when $\sigma_n \succeq a_n$ i.e. $\sigma_n \succeq e^{-n}$.

Similarly, if $\sigma_n \prec \exp\{-n\}$, setting $a_n \asymp \exp\{-n\}$ easily gives us an exponentially small lower bound.

In the passive setting, Eq.(\ref{active}) does not apply. Since the two convolved distributions differ only on an interval of size $2(\sigma_n+a_n)$, the effective number of points falling in this interval would be $\asymp n(\sigma_n+a_n)$.

When $\sigma_n \succeq a_n$, a simple calculation shows $$KL(P^0,P^1) ~\preceq~ n(\sigma_n+a_n) \frac{a_n^2}{\sigma_n^2} ~\asymp~ n\frac{a_n^2}{\sigma_n},$$
giving rise to a choice of $a_n \asymp \sqrt{\frac{\sigma_n}{n}}$, which is the passive minimax rate when $\sigma_n \succeq a_n$ i.e. $\sigma_n  \succeq \frac1{n}$.

When $\sigma_n \prec \frac1{n}$, a similar calculation shows
$$KL(P^0,P^1) ~\preceq~ n(\sigma_n+a_n) 4c^2 ~\asymp~ n a_n$$
giving rise to a choice of $a_n \asymp \frac1{n}$, which is the passive minimax rate when $\sigma_n \succeq a_n$ i.e. $\sigma_n  \prec \frac1{n}$.
\hfill $\blacksquare$

\paragraph{Proof of Theorem 2, $k>1$} We follow a very similar setup to the case $k=1$. The difference will lie in picking functions that are in $\mathcal{P}(c,C,k,\sigma_n)$ for general $k \neq 1$, and calculating the bounds on KL divergence appropriately. However, for notational convenience, we will assume that the domain is shifted to $[-\sigma_n,2-\sigma_n]$ instead of $[-1,1]$ and that the distance between thresholds is $a_n$ instead of $2a_n$. Define
\[
P_0(Y|x) = \begin{cases} 1/2 - c|x|^{k-1} \mbox{ if $x \in [-\sigma_n,0]$}\\ 1/2 + c|x|^{k-1} \mbox{ if $x > 0$}
\end{cases}
\]
\[
P_1(Y|x) = \begin{cases} 1/2 - c|x-a_n|^{k-1} \mbox{ if $x \in [-\sigma_n,a_n]$} \\
1/2 + c|x-a_n|^{k-1} \mbox{ if $x \in [a_n, \beta a_n+\sigma_n]$}\\
1/2 + c|x|^{k-1} \mbox{\ \ \ \ \ \ ~ if $x > \beta a_n+\sigma_n$}
\end{cases}
\]
where $\beta = \frac1{1-(c/C)^{1/(k-1)}} \geq 1$ is a constant chosen such that $P_1 \in \mathcal{P}(c,C,k,\sigma_n)$ (this fact is verified explicitly in the Appendix). For ease of notation, $P_0,P_1$ are  understood to actually saturate at $0,1$ if need be (i.e. we are implicitly working with $\min\{P_{0/1},1\}$, etc). The two thresholds are clearly at $0,a_n$ respectively, and after the point $\beta a_n+\sigma_n$, the two functions are the same. Continuing the same notation as for $k=1$, we let $P^i=P_i(Y|W)=F_i(w)$ for $i=0,1$.

The following claims hold true (Appendix).
\begin{enumerate}
\item When $\sigma_n \preceq a_n$, $\max_w|F_1(w) - F_2(w)| \asymp a_n^{k-1}$.
\item When $\sigma_n \succeq a_n$, $\max_w|F_1(w) - F_2(w)| \asymp \sigma_n^{k-2} a_n$.
\item As a subpart of the above cases, when $\sigma_n \asymp a_n$, $\max_w|F_1(w) - F_2(w)| \asymp \sigma_n^{k-2} a_n \asymp a_n^{k-1}$
\end{enumerate}

If the above propositions are true, we can verify:
\begin{enumerate}
\item In the first case, $KL(P^0,P^1) \preceq na_n^{2k-2}$, hence $a_n \asymp n^{-\frac1{2k-2}}$ is a lower bound when $\sigma_n \preceq n^{-\frac1{2k-2}}$.
\item Otherwise, $KL(P^0,P^1) \preceq n\sigma_n^{2k-4}a_n^2$, hence $a_n \asymp \frac{\sigma_n^{-(k-2)}}{\sqrt n}$ is a lower bound when $\sigma_n \succ n^{-\frac1{2k-2}}$.
\end{enumerate}

The passive bounds follow by not just considering the maximum difference between $|F_1(w) - F_2(w)|$ but also the length of that difference, since it is directly proportional to the number of points that may randomly fall in that region. Following the same calculations,
\begin{enumerate}
\item When $\sigma_n \prec a_n$, $|F_1(w) - F_2(w)| \asymp a_n^{k-1}$ for all $w\in[0,\beta a_n+2\sigma_n]$. Hence $KL(P^0,P^1) \preceq n(\beta a_n + 2\sigma_n) a_n^{2k-2} \asymp na_n^{2k-1}$ and $a_n \asymp n^{-\frac1{2k-1}}$ is the minimax passive rate when $\sigma_n \prec n^{-\frac1{2k-1}}$.
\item When $\sigma_n \succ a_n$, $|F_1(w) - F_2(w)| \asymp \sigma_n^{k-2} a_n$ for all $w \in [0,\beta a_n + 2\sigma_n]$. Hence $KL(P^0,P^1) \preceq n(\beta a_n + 2\sigma_n)\sigma_n^{2k-4}a_n^2$ and $a_n \asymp \sigma_n^{-({k-\frac{3}{2}})}\sqrt{\frac{1}{n}}$ is the minimax passive rate when $\sigma_n \succ n^{-\frac1{2k-1}}$.
\end{enumerate}
as verified from the Appendix calculation. \hfill $\blacksquare$

\section{Upper Bounds}

For passive sampling, we present a modified histogram estimator, WIDEHIST, when the noise level $\sigma_n$ is larger than the noiseless minimax rate of $1/n$. Assume for simplicity that the $n$ sampled points on $[-1,1]$ are equally spaced to mimic a uniform distribution, lying at $\frac{(2j-1)}{2n}$,  $j=1,...,n$.

\textbf{Algorithm WIDEHIST.} 
\begin{enumerate}
\item Divide $[-1,1]$ into $m$ bins of width $h > \frac{2}{n}$ so $m=\frac{2}{h} < n$. 
The $i^{\rm th}$ bin covers
$[-1+(i-1)h, -1+ih]$, $i \in \{1,...,m\}$ and hence each bin has $\frac{nh}{2}$ points.
Let $b_i$ be the average number of positive labels in bin $i$ of these $\frac{nh}{2}$ points.
\item Let $\hat p_i$ be the average of the $b_i$'s over a all bins within $\pm \sigma_n/2$ of bin $i$. We ``classify'' regions with $\hat p_i < 1/2$ as being $-$ and $\hat p_i > 1/2$ as being $+$, and return $\hat t$ as the center of the first bin from left to right where $\hat p_i$ crosses half.
\end{enumerate}

Observe that we need not operate on $[-1,1]$ with $n$ queries - WIDEHIST(D,B) could take as inputs any domain $D$ and any query budget $B$. The argument below hinges on the fact that the convolved regression function behaves linearly around $t$.

\paragraph{Proof of Theorem 3, $k=1$, (Passive).}
Let $i^*\in\{1,...,m\}$ denote the true bin $[(i^*-1)h,i^*h]$ that contains $t$. Let $\hat t$ be from bin $\hat i$, i.e.
 $\hat p_{\hat i} < 1/2$ and $\hat p_{\hat i+1} > 1/2$. We will argue that $\hat i$ is very close to $i^*$, in which case the point error we suffer is $|\hat i - i^*|h$. Specifically, we prove that all bins except $I^* =\{ i^* - 1, i^*, i^* + 1 \}$ will be ``classified'' correctly with high probability. In other words, we claim w.h.p. $\hat p_i < 1/2$ if $i < i^* - 1$ and $\hat p_i > 1/2$ if $i > i^*+1$.

Indeed, we can show (Appendix)\vspace{-0.05in}
\begin{eqnarray}
\mbox{For } i>i^*+2, \ \E[\hat p_{i}] \geq \E[\hat{p}_{i^*+2}] \geq 1/2 + \lbs h \label{pbt}\\
\mbox{For } i< i^*-2, \ \E[\hat{p}_{i}] \leq \E[\hat{p}_{i^*-2}] \leq 1/2 - \lbs h \label{pat}
\end{eqnarray}
Using Hoeffding's inequality, we get that for bin $i$,
$\Pr (|\hat{p}_i - p_i| > \epsilon) ~\leq~ 2\exp\left\{-2\tfrac{n\sigma_n}{2}\epsilon^2\right\}$
Taking union bound over all bins other than those in $i^*-1,i^*,i^*+1$ and setting $\epsilon=\lbs h$, we get
\begin{equation*}\vspace{-0.05in}
\Pr(\forall i \backslash I^*, |\hat{p}_i - p_i| > \lbs h) ~\leq~ 2m \exp\left\{-2\tfrac{n\sigma_n}{2}\left(\tfrac{c h}{\sigma_n}\right)^2\right\}
\end{equation*}
So we get bins $i\backslash I^*$ correct and $\hat i \in \{i^*~-~1,i^*,i^*~+~1\}$ with probability $\geq 1 -2n \exp\left\{-n\sigma_n\left(\tfrac{c h}{\sigma_n}\right)^2\right\}$ since $m~<~n$. Setting $h = \frac1{c}\sqrt{\frac{\sigma_n}{n} \log(\tfrac{2n}{\delta})}$ makes this hold with probability $\geq 1-\delta$ so the point error $|\hi - i^*|h < 2h$ behaves like $h \preceq \sqrt{\frac{\sigma_n}{n}}$. \hfill $\blacksquare$

For active sampling when the noise level $\sigma_n$ is larger than the minimax noiseless rate $e^{-n}$, we present a algorithm ACTPASS which makes its $n$ queries on the domain $[-1,1]$ in $E$ different epochs/rounds. As a subroutine, it uses any optimal passive learning algorithm, like WIDEHIST(D,B). In each round, ACTPASS runs WIDEHIST on progressively smaller domains D with a restricted budget B. Hence it ``activizes'' the WIDEHIST and achieves the optimal active rate in the process. This algorithm was inspired by a similar idea from \cite{RS13}.

\textbf{Algorithm ACTPASS.} 

Let $E = \lceil \log(1/\sigma_n) \rceil$ be the number of epochs and $D_1=[-1,1]$ denote the domain of ``radius'' $R_1=1$ around $t_0=0$. The budget of every epoch is a constant $B = n/E$. For epochs $1 \leq e \leq E$, do:
\begin{enumerate}
\item Query for $B$ labels uniformly on $D_e$.
\item Let $t_e = {\rm WIDEHIST}(D_e,B)$ be the returned estimator using the most recent samples and labels. 
\item Define $D_{e+1} = [t_e - 2^{-e}, t_e + 2^{-e}] \cap [-1,1]$ with a radius of at most $R_{e+1}=2^{-e}$ around $t_e$. Repeat.
\end{enumerate}
Observe that ACTPASS runs while $R_e > \sigma_n$, since by design $E \geq \log(1/\sigma_n)$ so $\sigma_n \leq 2^{-E} = R_{E+1}$.

\paragraph{Proof of Theorem 3, $k=2$, (Active).}
The analysis of ACTPASS proceeds in two stages depending on the value of $\sigma_n$. Initially, when $R_e$ is large, it is possible that $\sigma_n \preceq R_e/n$ and in this phase, the passive algorithm WIDEHIST will behave as if it is in the noiseless setting since the noise is smaller than its noiseless rate. However, after some point, when $R_e$ becomes quite small, $\sigma_n \succeq R_e/n$ is possible and then WIDEHIST will behave as if it is in the noisy setting since noise is larger than its noiseless rate. Observe that it cannot stay in the first phase till the end of the algorithm, since the first phase runs while $\sigma_n \preceq R_e/n$ but we know that $\sigma_n > R_{E+1}$ by construction, so there must be an epoch where it switches phases, and ends the algorithm in its second phase. 

We prove (by a separate induction in each epoch) that with high probability, the true threshold $t$ will always lie inside the  domain at the start of every epoch (this is clearly true before the first epoch). We claim:
\begin{enumerate}
\item Before all $e$ in phase one, $t \in D_e$ w.h.p.
\item Before all $e$ in phase two, $t \in D_e$ w.h.p.
\end{enumerate}
We prove these in the Appendix. If these are true, then in the second phase, WIDEHIST is in the large noise setting and it gets an error of $\sqrt{\frac{R_e\sigma_n}{B}}$. Hence the final error of the algorithm is $\sqrt{\frac{R_E\sigma_n}{n/E}} \asymp \frac{\sigma_n}{\sqrt n}$.
\hfill $\blacksquare$

\paragraph{Proof of Theorem 3, $k>1$.} The proofs for $k>1$ are simply generalizations of those for $k=1$. Again, we present concise arguments here for the settings where the algorithm can actually detect noise, i.e. when the noise level is larger than the noiseless minimax rate (otherwise, one can argue that algorithms which worked for the noiseless case will suffice). In both cases, the algorithm remains unchanged.

1. We outline the proof for WIDEHIST when $\sigma_n \succeq n^{-\frac1{2k-1}}$. Using similar notation as before, we will again show that if $t$ is in bin $i^*$ of width $h < \sigma_n$, then except for bins $i^*-1,i^*,i^*+1$, we will "classify" all other bins correct with high probability, by averaging over the $n \sigma_n / 2$ points to the left and right of that bin. Specifically, we claim
\begin{eqnarray}
\mbox{For $i>i^*+2$, } \E[\hat p_{i}] \geq \E[\hat{p}_{i^*+2}] ~\geq~ 1/2 + \lambda \sigma_n^{k-2} h~ \label{pbt}\\
\mbox{For $i < i^*-2$, } \E[\hat{p}_{i}] \leq \E[\hat{p}_{i^*-2}] ~\leq~ 1/2 - \lambda \sigma_n^{k-2} h~ \label{pat}
\end{eqnarray}
A similar use of Hoeffding's inequality gives 
\begin{eqnarray*}
\Pr(\forall i \backslash I^*, |\hat{p}_i - p_i| &>& \lambda \sigma_n^{k-2} h) ~\leq~\\
&& 2m \exp \left\{-2 (\tfrac{n \sigma_n}{2R}) h^2 \lambda^2 \sigma_n^{2k-4}\right\}.
\end{eqnarray*}
Arguing as before, w.h.p. we get a point error of $h \preceq \sqrt {\frac{R}{\sigma_n^{2k-3} n}} < \sigma_n$ when $\sigma_n \succ n^{-\frac1{2k-1}}$.\\

2. We outline the proof for ACTPASS when $\sigma_n \succeq n^{-\frac1{2k-2}}$. As before, the algorithm runs in two phases, and we will prove required properties within each phase by induction. 

The first phase is when $R_e$ is large and so $\sigma_n$ may possibly be smaller than $(R_e/n)^{\frac1{2k-1}}$ and WIDEHIST will achieve noiseless rates within each epoch. In the second phase, after $R_e$ has shrunk enough, $\sigma_n$ will become larger than $(R_e/n)^{\frac1{2k-1}}$ and WIDEHIST will achieve noisy rates in these epochs. 

One can verify, as before, that the second phase must occur, by design. Intuitively, the second phase must occur because we make a fixed number of queries $n/E \asymp n/\log n$ in a halving domain size (equivalently we make geometrically increasing queries on a rescaled domain), and so relatively in successive epochs this noiseless error shrinks, and at some point $\sigma_n$ becomes larger than this shrinking noiseless error rate.

As before we make the following claims:
\begin{enumerate}
\item Before all $e$ in phase one $t \in D_e$ w.h.p.
\item Before all $e$ in phase two $t \in D_e$ w.h.p.
\end{enumerate}
These are proved in the Appendix by induction.

The final point error is given by WIDEHIST in the last epoch as  $\sqrt {\frac{R_E }{\sigma_n^{2k-3} n/E}} \asymp \frac1{\sigma_n^{k-2}} \sqrt {\frac1{ n}}$ since $R_E \asymp \sigma_n$ and $E \asymp \log n$.

\section{Conclusion}

In this paper, we propose a simple Berkson error model for one-dimensional threshold classification, inspired by the setup and model analysed in \cite{CN07,CN08}, in which we can analyse active learning with additive uniform feature noise. To the best of our knowledge, this is the first attempt at jointly tackling feature noise and label noise in active learning. 

This simple setting already yields interesting behaviour depending on the additive feature noise level and the label noise of the underlying regression function. For both passive and active learning, whenever the noise level is smaller than the minimax noiseless rate, the learner cannot notice that there is noise, and will continue to achieve the noiseless rate. As the noise gets larger, the rates do depend on the noise level. Importantly, one can achieve better rates than passive learning in most scenarios, and we propose unique algorithms/estimators to achieve tight rates. The idea of ``activizing'' passive algorithms, like algorithm ACTPASS did, seems especially powerful and could carry forward to other settings beyond our paper and \cite{RS13}. 

The immediate future work and most direct extension to this paper concerns the main weakness of the paper - the possibility of getting rid of Assumption (M), which is the only hurdle to a fair comparision with the noiseless setting. We would like to re-emphasize that at first glance, the rates may be  misleading and counterintuitive because it ``appears'' as if larger noise could possibly help estimation due to the presence of $\sigma_n$ in the denominator for larger $k$. 

However, we point out once more that the class of functions is not constant over all $\sigma_n$ - it depends on $\sigma_n$, and in fact it gets ``smaller'' in some sense with larger $\sigma_n$ because the assumption (M) becomes more stringent. This observation about the non-constant function class, along with the fact that convolution with uniform noise seems to \textit{unflatten} the regression function as shown in the figures, together cause the rates to seemingly improve with larger noise levels.

Analysing the case without (M) seems to be quite a challenging task since the noiseless and convolved thresholds can be different - we did attempt to formulate a few kernel-based estimators with additional assumptions, but do not presently have tight bounds, and leave those for a future work.

\subsection*{Acknowledgements}
We thank Rui Castro for detailed conversations about our model and results. This work is supported in part by NSF Big Data grant IIS-1247658.

\bibliographystyle{natbib}
\bibliography{biblio}

\newpage

\appendix

\section{Justifying Claims in the Lower Bounds}


Approximations:
\begin{enumerate}
\item $(x+y)^k = x^k(1+y/x)^k \approx x^k + kx^{k-1}y$ when $y \prec x$. Even when $y \preceq x$, both terms are the same order.
\item  $(x-y)^k = x^k(1-y/x)^k \approx x^k - kx^{k-1}y$ when $y \prec x$. Even when $y \preceq x$ both terms are the same order.
\item When $y < x$ but not $y \prec x$, by Taylor expansion of $(1+z)^k$ around $z=0$, we have $(x+y)^k = x^k(1+y/x)^k = x^k[1 + (1+c)^{k-1}y/x] = x^k + Cx^{k-1}y$ for some $0 < c < y/x < 1$ and some constant $C$. Similarly for $(x-y)^k$.
\end{enumerate}

Let's assume the boundary is at $-\sigma$ for easier calculations. (we denote $a_n,\sigma_n$ as $a,\sigma$ here). Remember
\[
m_1(x) =  1/2 + cx|x|^{k-2} \mbox{ if } x\geq -\sigma
\]
\[
m_2(x) = \begin{cases} 1/2 + c(x-a)|x-a|^{k-2}& \mbox{ if }  x < \beta a+\sigma\\
m_1(x)& \mbox{ if } x \geq \beta a+\sigma
\end{cases}
\]
where $\beta = \frac1{1-(c/C)^{1/(k-1)}} \geq 1$ is such that $m_2 \in P(\kappa,c,C,\sigma)$. Clearly, when $x < \beta a + \sigma$, $m_2$ satisfies condition (T). So, we
only need to verify that whenever $x \geq \beta a + \sigma$ we  have
\begin{eqnarray}
m_2(x) -1/2 ~=~ cx^{k-1} &\leq& C(x-a)^{k-1}
\end{eqnarray}
This statement holds $
\mbox{ iff ~}~ (c/C)^{1/(k-1)} \leq 1 - a/x
~ \Leftrightarrow ~ a/x \leq 1-(c/C)^{1/(k-1)}
~ \Leftrightarrow ~ x \geq \beta a$, 
which holds for all $\sigma \geq 0$, and hence $m_2$ satisfies condition (T).

Proposition 1. When $\sigma \prec a$, $\max_w |F_1(w) - F_2(w)| \asymp a^{k-1}$\\
Proposition 2. When $\sigma \succ a$ $\max_w |F_1(w) - F_2(w)| \asymp \sigma^{k-2}a$

Let us now prove these two propositions, with detailed calculations in each case (note that when $\sigma \asymp a$, then $\max_w |F_1(w) - F_2(w)| \asymp a^{k-1} \asymp \sigma^{k-2}a$, and can be checked using our approximations 1,2,3).


\begin{enumerate}
\item When $\sigma \prec a$, we will prove proposition 1. Remember that we can't query in $-\sigma \leq w \leq 0$.
\begin{enumerate}
\item
When $0\leq w \leq  \sigma$, we have
\begin{eqnarray*}
F_1(w) = (m_1 \star U )(w) &=& \int_{w-\sigma}^0 (1/2 -  cx|x|^{k-2}) dx/{2\sigma} + \int_0^{w+\sigma} (1/2 +  cx^{k-1}) dx/2\sigma\\
&=& 1/2 + \frac{c}{2\sigma k}[(w+\sigma)^k - ( \sigma-w)^k]\\
&=& 1/2 + \frac{c}{2\sigma k}\sigma^k[(1+w/\sigma)^k - (1-w/\sigma)^k]\\
&\approx&  1/2 + c \sigma^{k-2}w 
\end{eqnarray*}
\begin{eqnarray*}
F_2(w) = (m_2 \star U )(w) &=& \int_{w-\sigma}^{w+\sigma} (1/2 -  c(x-a)|x-a|^{k-2}) dx/{2\sigma} \\
&=& 1/2 - \frac{c}{2\sigma k}[(a - w - \sigma)^k - (a+\sigma - w)^k]\\
&\approx& 1/2 - c(a-w)^{k-1}\\
\end{eqnarray*}
[Boundaries: $F_1(0) - \frac1{2}=0, F_1(\sigma) -\frac1{2} \asymp \sigma^{k-1},F_2(0) -\frac1{2} \asymp -a^{k-1}, F_2(\sigma)-\frac1{2} \asymp -a^{k-1} $]. 
\begin{eqnarray*}
F_1(w) - F_2(w) &\preceq& a^{k-1}
\end{eqnarray*}

\item When $\sigma \leq w \leq  a-\sigma$
\begin{eqnarray*}
F_1(w) = (m_1 \star U )(w) &=& \int_{w-\sigma}^{w+\sigma} (1/2 +  cx^{k-1}) dx/{2\sigma} \\
&=& 1/2 + \frac{c}{2\sigma k}[(w+\sigma)^k - (w- \sigma)^k]\\
&\approx& 1/2 + cw^{k-1}
\end{eqnarray*}
\begin{eqnarray*}
F_2(w) = (m_2 \star U )(w) &=& \int_{w-\sigma}^{w+\sigma} (1/2 -  c(x-a)|x-a|^{k-2}) dx/{2\sigma} \\
&=& 1/2 - \frac{c}{2\sigma k}[(a - w - \sigma)^k - (a+\sigma - w)^k]\\
&\approx& 1/2 - c(a-w)^{k-1}
\end{eqnarray*}
[Boundaries: $F_1(\sigma)-\frac1{2}\asymp \sigma^{k-1}, F_1(a-\sigma)-\frac1{2}\asymp a^{k-1}, F_2(\sigma)-\frac1{2}\asymp -a^{k-1}, F_2(a-\sigma)-\frac1{2}\asymp -\sigma^{k-1}$]. 
\begin{eqnarray*}
F_1(w) - F_2(w) &=& cw^{k-1} + c(a-w)^{k-1} \\
&\leq& c(a - \sigma)^{k-1} + c (a - \sigma)^{k-1} \\
&\preceq& a^{k-1}
\end{eqnarray*}

\item When $a - \sigma \leq w \leq a$
\begin{eqnarray*}
F_1(w) 
&\approx& 1/2 + cw^{k-1}
\end{eqnarray*}
\begin{eqnarray*}
F_2(w)  &=& \int_{w-\sigma}^{a} (1/2 -  c(x-a)|x-a|^{k-2}) dx/{2\sigma} + \int_{a}^{w+\sigma} 1/2 + c(x-a)^{k-1} dx/{2\sigma}\\
&=& 1/2 - \frac{c}{2\sigma k}[(a-w+\sigma)^k - (w+\sigma - a)^k]\\
&\approx& 1/2 - c\sigma^{k-2}(a-w)
\end{eqnarray*}
[Boundaries: $F_1(a-\sigma)-\frac1{2} \asymp a^{k-1}, F_1(a)-\frac1{2}\asymp a^{k-1}, F_2(a-\sigma)-\frac1{2}\asymp -\sigma^{k-1},F_2(a)-\frac1{2}=0$]
\begin{eqnarray*}
F_1(w) - F_2(w) &\approx& cw^{k-1} + c\sigma^{k-2}(a-w) \\
&\leq& ca^{k-1} + c \sigma^{k-2} \sigma\\
&\preceq& a^{k-1}
\end{eqnarray*}

\item When $a \leq w \leq a+\sigma$
\begin{eqnarray*}
F_1(w) 
&\approx& 1/2 + cw^{k-1}
\end{eqnarray*}
\begin{eqnarray*}
F_2(w)  &\approx& 1/2 +  c\sigma^{k-2}(a-w)
\end{eqnarray*}
[Boundaries: $F_1(a)-\frac1{2}\asymp a^{k-1},F_1(a+\sigma)-\frac1{2}\asymp a^{k-1}, F_2(a)-\frac1{2}=0, F_2(a+\sigma)-\frac1{2}\asymp \sigma^{k-1}$]
$$
F_1(w) - F_2(w) \preceq a^{k-1}
$$

\item When $a + \sigma \leq w \leq \beta a-\sigma$
\begin{eqnarray*}
F_1(w) 
&\approx& 1/2 + cw^{k-1}
\end{eqnarray*}
\begin{eqnarray*}
F_2(w)  &=&  \int_{w-\sigma}^{w+\sigma} 1/2 + c(x-a)^{k-1} dx/{2\sigma}\\
&=& 1/2 + \frac{c}{2\sigma k}[(w+\sigma - a)^k - (w-\sigma-a)^k]\\
&\approx& 1/2 + c(w-a)^{k-1}
\end{eqnarray*}
[B: $F_1(a+\sigma)-\frac1{2}\asymp a^{k-1}, F_1(\beta a - \sigma) - \frac1{2}\asymp a^{k-1}, F_2(a+\sigma)-\frac1{2}\asymp \sigma^{k-1}, F_2(\beta a -\sigma) - \frac1{2} \asymp a^{k-1}$]
\begin{eqnarray*}
F_1(w) - F_2(w) &\approx& cw^{k-1} - c(w-a)^{k-1}\\
&\leq& c(\beta a - \sigma)^{k-1} + c\sigma^{k-1}\\
&\leq& c(\beta^{k-1}+1)a^{k-1}\\
&\preceq& a^{k-1}
\end{eqnarray*}

\item When $\beta a - \sigma \leq w \leq \beta a + \sigma$
\begin{eqnarray*}
F_1(w) 
&\approx& 1/2 + cw^{k-1}
\end{eqnarray*}
\begin{eqnarray*}
F_2(w)  &=&  \int_{w-\sigma}^{\beta a} 1/2 + c(x-a)^{k-1} dx/{2\sigma} + \int_{\beta a}^{w+\sigma}1/2 + x^{k-1}dx/2\sigma\\
&=& 1/2 +\frac{c}{2\sigma k}[(\beta a - a)^k - (w-\sigma-a)^k + (w+\sigma)^k - (\beta a)^{k} ]
\end{eqnarray*}
[$F_1(\beta a - \sigma) - \frac1{2}\asymp a^{k-1}, F_1(\beta a+\sigma)-\frac1{2}\asymp a^{k-1}, F_2(\beta a -\sigma) - \frac1{2} \asymp a^{k-1}, F_2(\beta a+\sigma)-\frac1{2}\asymp a^{k-1}$]
\begin{eqnarray*}
F_1(w) - F_2(w) &=& cw^{k-1} + \frac{c}{2\sigma k}[(\beta^k - (\beta-1)^k)a^k + (w-\sigma-a)^k - (w-\sigma)^k]\\
&\leq& c(\beta+1)^{k-1}a^{k-1} + \frac{c}{2\sigma k}[(\beta a)^k - (\beta a - 2\sigma)^k] -\frac{c}{2\sigma k}[(\beta-1)^k a^k - ((\beta-1)a - \sigma)^k]\\
&\approx& c(\beta+1)^{k-1}a^{k-1} + \frac{c}{2\sigma k}[k(\beta a)^{k-1} 2\sigma] -\frac{c}{2\sigma k}[k(\beta-1)^{k-1}a^{k-1} \sigma]\\
&=& c a^{k-1} [(\beta+1)^{k-1} + \beta^{k-1} - \tfrac1{2}(\beta-1)^{k-1}]\\
&\asymp& a^{k-1}
\end{eqnarray*}

\item When $\beta a + \sigma \leq w \leq \beta a + 2\sigma$
\begin{eqnarray*}
F_1(w) 
&=& 1/2 + \frac{c}{2\sigma k}[(w+\sigma)^k - (w-\sigma)^k]
\end{eqnarray*}
\begin{eqnarray*}
F_2(w) &=& \int_{w-\sigma}^{\beta a + \sigma} 1/2 + c(x-a)^{k-1} dx/2\sigma + \int_{\beta a +\sigma}^{w+\sigma} 1/2 + cx^{k-1} dx/2\sigma \\
&=& 1/2 + \frac{c}{2\sigma k} [ (\beta a + \sigma - a)^k - (w- \sigma-a)^k + (w+\sigma)^k - (\beta a + \sigma)^k ] \\
\end{eqnarray*}
[$F_1(\beta a+\sigma)-\frac1{2}\asymp a^{k-1}, F_1(\beta a + 2\sigma) - \frac1{2}\asymp a^{k-1},  F_2(\beta a+\sigma)-\frac1{2}\asymp a^{k-1}, F_2(\beta a +2\sigma) - \frac1{2} \asymp a^{k-1}$]
\begin{eqnarray*}
F_1(w) - F_2(w) &=& \frac{c}{2\sigma k} [(\beta a + \sigma)^k - (\beta a + \sigma - a)^k +(w-\sigma-a)^k-(w-\sigma)^k]\\
&\approx& \frac{c}{2\sigma k} [(\beta a + \sigma)^{k-1} ka - (w-\sigma)^{k-1}ka]\\
&\leq&  \frac{ca}{2\sigma} [(\beta a + \sigma)^{k-1} - (\beta a)^{k-1}]\\
&\approx& \frac{ca}{2\sigma} [(\beta a)^{k-1} (1+\frac{(k-1)\sigma}{\beta a}) - (\beta a)^{k-1}]\\
&=& a^{k-1} [c\beta^{k-2}(k-1)/2]\\
&\asymp& a^{k-1}
\end{eqnarray*}

\item When $w \geq \beta a + 2\sigma$
$$
F_1(w) = F_2(w)
$$


\end{enumerate}

That completes the proof of the first claim. 

\item When $\sigma \succ a$, we will prove the second proposition.
\begin{enumerate}
 \item
When $-\sigma \leq w \leq 0$, we are not allowed to query here.

\item When $0 < w \leq \beta a$
\begin{eqnarray*}
F_1(w) = (m_1 \star U )(w) &=& \int_{w-\sigma}^0 (1/2 -  cx|x|^{k-2}) dx/{2\sigma} + \int_0^{w+\sigma} (1/2 +  cx^{k-1}) dx/2\sigma\\
&=& 1/2 + \frac{c}{2\sigma k}[(w+\sigma)^k - ( \sigma-w)^k]\\
&=& 1/2 + \frac{c}{2\sigma k}\sigma^k[(1+w/\sigma)^k - (1-w/\sigma)^k]\\
&\approx&  1/2 + c \sigma^{k-2}w 
\end{eqnarray*}

Similarly $F_2(w) \approx 1/2 + c\sigma^{k-2}(w-a)$ 

[Boundaries: $F_1(0)-\frac1{2}=0, F_1(\beta a)-\frac1{2} \asymp \sigma^{k-2}a, F_2(0)-\frac1{2}\asymp -\sigma^{k-2}a, F_2(\beta a)\asymp \sigma^{k-2}a$]
$$F_1(w) - F_2(w) \asymp \sigma^{k-2}_n a.$$

\item When $\beta a \leq w \leq \sigma$
\begin{eqnarray*}
F_1(w) = &=& \int_{w-\sigma}^0 (1/2 -  cx|x|^{k-2}) dx/{2\sigma} + \int_0^{w+\sigma} (1/2 +  cx^{k-1}) dx/2\sigma\\
&=& 1/2 + \frac{c}{2\sigma k}[(w+\sigma)^k - ( \sigma-w)^k]\\
&=& 1/2 + \frac{c}{2\sigma k}\sigma^k[(1+w/\sigma)^k - (1-w/\sigma)^k]\\
&\approx&  1/2 + c \sigma^{k-2}w 
\end{eqnarray*}
\begin{eqnarray*}
F_2(w) &=& \int_{w-\sigma}^{a} (1/2 - c(x-a)|x-a|^{k-2})\frac{dx}{2\sigma} + \int_{a}^{\beta a + \sigma} (1/2 + c(x-a)^{k-1}) \frac{dx}{2\sigma} + \int_{\beta a + \sigma}^{w+\sigma} 1/2 + cx^{k-1} \frac{dx}{2\sigma}\\
&=& 1/2 + \frac{c}{2\sigma k} [-(\sigma + a - w)^k + (\beta a + \sigma-a)^k + (w+\sigma)^k - (\beta a + \sigma)^k]\\
&\approx& 1/2 + \frac{c}{2\sigma k} [ -\sigma^k(1-\frac{k(w-a)}{\sigma}) + \sigma^k (1 + \frac{k(\beta-1)a}{\sigma}) + \sigma^k (1 + \frac{kw}{\sigma}) - \sigma^k (1 + \frac{k\beta a}{\sigma})]\\
&=& 1/2 + \frac{c}{2} \sigma^{k-2} [w-a + (\beta-1) a + w - \beta a]\\
&=& 1/2 + c\sigma^{k-2}(w-a)
\end{eqnarray*}
[Boundaries: $F_1(\beta a)-\frac1{2} \asymp \sigma^{k-2}a, F_1(\sigma)-\frac1{2}\asymp \sigma^{k-1}, F_2(\beta a)\asymp \sigma^{k-2}a, F_2(\sigma)-\frac1{2}\asymp -\sigma^{k-2}a$]
$$
F_1(w) - F_2(w) \asymp \sigma^{k-2}a
$$
Specifically, verify the boundary at $\sigma$

\begin{eqnarray*}
F_1(\sigma)-F_2(\sigma) &=& \frac{c}{2\sigma k}[a^k - (\beta a + \sigma - a)^k + (\beta a + \sigma)^k]\\
&=& \frac{c}{2\sigma k}[a^k - \sigma^k(1+k\frac{\beta a - a}{\sigma}) + \sigma^k(1 + k\frac{\beta a}{\sigma}) ]\\
&=& \frac{c}{2\sigma k}[a^k + k\sigma^{k-1}a]\\
&\leq& c\sigma^{k-2}a
\end{eqnarray*}

\item When $ \sigma \leq w \leq  a + \sigma$
\begin{eqnarray*}
F_1(w) &=& \int_{w-\sigma}^{w+\sigma}(1/2+cx^{k-1})dx/2\sigma\\
&=&1/2 + \frac{c}{2\sigma k}[(w+\sigma)^k - (w-\sigma)^k]\\
\end{eqnarray*}
\begin{eqnarray*}
F_2(w) &=& \int_{w-\sigma}^{a} (1/2 - c(x-a)|x-a|^{k-2})\frac{dx}{2\sigma} + \int_{a}^{\beta a + \sigma} (1/2 + c(x-a)^{k-1}) \frac{dx}{2\sigma} + \int_{\beta a + \sigma}^{w+\sigma} 1/2 + cx^{k-1} \frac{dx}{2\sigma}\\
&=& 1/2 + \frac{c}{2\sigma k} [-(\sigma + a - w)^k + (\beta a + \sigma-a)^k + (w+\sigma)^k - (\beta a + \sigma)^k]
\end{eqnarray*}
\begin{eqnarray*}
F_1(w) - F_2(w) &=& \frac{c}{2\sigma k} [(\sigma + a - w)^k - (\beta a + \sigma-a)^k - (w-\sigma)^k + (\beta a + \sigma)^k]
\end{eqnarray*}
Differentiating the above term with respect to $w$, gives $\frac{c}{2\sigma } [-(\sigma + a - w)^{k-1} - (w-\sigma)^{k-1}] \leq 0$ because $\sigma \leq w \leq a + \sigma$ and hence $F_1(w) - F_2(w)$ is decreasing with $w$.
We already saw $F_1(\sigma) - F_2(\sigma) \leq c\sigma^{k-2}a$. We can also verify that at the other boundary,
\begin{eqnarray*}
F_1(a+\sigma) - F_2(a+\sigma) &=& \frac{c}{2\sigma k} [- (\beta a + \sigma-a)^k - a^k + (\beta a + \sigma)^k]\\
&=& \frac{c}{2\sigma k}[-a^k - \sigma^k(1+k\frac{\beta a - a}{\sigma}) + \sigma^k(1 + k\frac{\beta a}{\sigma}) ]\\
&=& \frac{c}{2\sigma k}[-a^k + k\sigma^{k-1}a]\\
&\leq& \frac{c}{2}\sigma^{k-2}a
\end{eqnarray*}

\item When $ \sigma + a \leq w \leq \beta a + \sigma$
\begin{eqnarray*}
F_1(w) &=& \int_{w-\sigma}^{w+\sigma}(1/2+cx^{k-1})dx/2\sigma\\
&=&1/2 + \frac{c}{2\sigma k}[(w+\sigma)^k - (w-\sigma)^k]\\
\end{eqnarray*}
\begin{eqnarray*}
F_2(w) &=& \int_{w-\sigma}^{\beta a + \sigma} (1/2 + c(x-a)^{k-1}) \frac{dx}{2\sigma} + \int_{\beta a + \sigma}^{w+\sigma} 1/2 + cx^{k-1} \frac{dx}{2\sigma}\\
&=& 1/2 + \frac{c}{2\sigma k} [ (\beta a + \sigma-a)^k -(w-\sigma - a)^k + (w+\sigma)^k - (\beta a + \sigma)^k]\\
\end{eqnarray*}
\begin{eqnarray*}
F_1(w) - F_2(w) &=& \frac{c}{2\sigma k} [(w-\sigma - a)^k - (\beta a + \sigma-a)^k - (w-\sigma)^k + (\beta a + \sigma)^k]\\
\end{eqnarray*}
Differentiating with respect to $w$ gives $\frac{c}{2\sigma}[(w-\sigma - a)^{k-1} - (w-\sigma)^{k-1}] \leq 0$ because $w - \sigma - a \leq w-\sigma$ and so $F_1-F_2$ is decreasing with $w$. We  know $F_1(a+\sigma) - F_2(a+\sigma) \leq \frac{c}{2}\sigma^{k-2}a$, and we can verify at the other boundary that 
\begin{eqnarray*}
F_1(\beta a+\sigma) - F_2(\beta a+\sigma) &=& \frac{c}{2\sigma k} [(\beta a - a)^k - (\beta a + \sigma-a)^k - (\beta a)^k + (\beta a + \sigma)^k]\\
&\approx& \frac{c}{2\sigma k} [(\beta a - a)^k - (\beta a)^k - \sigma^k(1 + k\frac{\beta a-a }{\sigma}) + \sigma^k(1 + k\frac{\beta a}{\sigma})]\\
&=&  \frac{c}{2\sigma k} [ (\beta a - a)^k - (\beta a)^k + k\sigma^{k-1}a ]\\
&\leq& \frac{c}{2}\sigma^{k-2}a
\end{eqnarray*}

\item When $\beta a + \sigma \leq w \leq \beta a + 2\sigma$
$$
F_1(w) = 1/2 + \frac{c}{2\sigma k}[(w+\sigma)^k - ( w-\sigma)^k]
$$
\begin{eqnarray*}
F_2(w) &=& \int_{w-\sigma}^{\beta a + \sigma} 1/2 + c(x-a)^{k-1}dx/2\sigma + \int_{\beta a + \sigma}^{w+\sigma} 1/2 + cx^{k-1}dx/2\sigma \\ 
&=&1/2 + \frac{c}{2k\sigma} [(\beta a + \sigma - a)^k - (w-\sigma - a)^k + (w+\sigma)^k - (\beta a + \sigma)^k]
\end{eqnarray*}
Hence
\begin{eqnarray*}
F_1(w) - F_2(w) &=& \frac{c}{2\sigma k} [(\beta a + \sigma)^k - (\beta a + \sigma - a)^k +(w-\sigma-a)^k-(w-\sigma)^k]\\
&\approx& \frac{c}{2\sigma k} [(\beta a + \sigma)^{k-1} ka - (w-\sigma)^{k-1}ka]\\
&\leq&  \frac{ca}{2\sigma} [(\beta a + \sigma)^{k-1} - (\beta a)^{k-1}]\\
&\approx& c/2 \sigma^{k-2} a \\
&\asymp& \sigma^{k-2}a
\end{eqnarray*}
Alternately, by the same argument as in the previous case, differentiating with respect to $w$ gives $\frac{c}{2\sigma}[(w-\sigma - a)^{k-1} - (w-\sigma)^{k-1}] \leq 0$ because $w - \sigma - a \leq w-\sigma$ and so $F_1-F_2$ is decreasing with $w$. We  know $F_1(\beta a+\sigma) - F_2(\beta a+\sigma) \leq \frac{c}{2}\sigma^{k-2}a$, and we can verify at the other endpoint that 
\begin{eqnarray*}
F_1(\beta a+2\sigma) - F_2(\beta a+2\sigma) &=& 0
\end{eqnarray*}

\item When $w \geq \beta a + 2\sigma$, 
$F_1(w) = F_2(w)$

\end{enumerate}
That completes the proof of the second proposition.
\end{enumerate}


\section{Convolved Regression Function, Justifying Eqs.(8-11)}

For ease of presentation, let us assume the threshold is at 0, and define $m \in \mathcal{P}(c,C,k,\sigma) $ as 
$$
m(x) = \begin{cases} 1/2 + f(x) + \Delta(x) ~\mbox{ if $x \geq 0$}\\ 1/2 - f(x) ~\mbox{if $x < 0$} \end{cases}
$$
Due to assumption (M), $\Delta(x)$ must be $0$ when $0\leq x \leq \sigma$. Hence, the Taylor expansion of $\Delta(x)$ around $x=\sigma$ looks like 
$$
\Delta(x) = (x-\sigma)\Delta'(\sigma) + (x-\sigma)^2 \Delta''(\sigma) + ...
$$

If one represents, as before, $F(x) = m \star U$, then directly from the definitions, it follows for $\delta > 0$ that
$$
F(\delta) - F(0) = \int_{\sigma}^{\sigma+\delta} (1/2 + f(z) + \Delta(z)) \frac{dz}{2\sigma} - \int_{-\sigma}^{-\sigma+\delta} (1/2 - f(z)) \frac{dz}{2\sigma}
$$
In particular, due to the form (T) of $m$, let $f = c_1|x|^{k-1}$ for some $c \leq c_1 \leq C$ (we could also break $f$ into parts where it has different $c_1$s but this is a technicality and does not change the behaviour). Then
\begin{eqnarray*}
F(\delta) - F(0) &=& \frac{c_1}{2k\sigma}[ (x^k)_\sigma^{\sigma+\delta} - (x^k)_{-\sigma}^{-\sigma+\delta}] + \int_\sigma^{\delta+\sigma} [(z-\sigma)\Delta'(\sigma) + (z-\sigma)^2 \Delta''(\sigma) + ...] \frac{dz}{2\sigma}\\
&=& \frac{c_1}{2k\sigma}[(\sigma + \delta)^k - \sigma^k + (-\sigma + \delta)^k - (-\sigma)^k] + \frac{[(z-\sigma)^2]_{\sigma}^{\sigma+\delta}}{4\sigma}\Delta'(\sigma) + ...\\
&\approx& c_1 \sigma^{k-2} \delta + \frac{\delta^2}{4\sigma}\Delta'(\sigma) + o(\delta^2)
\end{eqnarray*}

Thus we get  behaviour of the form 
$$F(t+h) \geq  1/2 + c\sigma^{k-2}h $$
One can derive similar results when $\delta < 0$.

The claims about WIDEHIST immediately follow from the above, but we can make them a little more explicit.
First note that $F(w) = 1/2 + \lbs (w-t)$ for $w$ close to $t$ (in fact for $w \in [t-\sigma,t+\sigma]$), as seen in Section 1 of this Appendix. Consider a bin just outside the bins $i^*-1,i^*,i^*+1$, for instance bin $i = i^*+2$ centered at $b_i$ (note $b_i \geq t+h$), and let $J$ be the set of points $j$ that fall within $b_i \pm \sigma/2$. Define 
$$
\hat p_i = \frac1{n\sigma/2R} \sum_{j \in J} \mathbb{I} (Y_j = +)
$$
where $Y_j \in \{\pm 1\}$ are observations at points $j \in J$. Now, we have, since $P (Y_j=+ ) = F(j)$
\begin{eqnarray*}
\E[\hat p_i]  &=&  \frac1{n\sigma/2R} \sum_{j \in J} F(j)\\
&=&  \frac1{n\sigma/2R} \left[ \sum_{j \in J} 1/2 + \lbs (X_j - t) \right]\\
&\approx& 1/2 + \frac1{\sigma} \int_{b_i-t-\sigma/2}^{b_i - t + \sigma/2} \lbs z dz \\
&=& 1/2 + \frac{c}{2\sigma^2} \left[(b_i - t + \sigma/2)^2 - (b_i -t - \sigma/2)^2 \right] \\
&=& 1/2 +  \lbs (b_i - t)\\
&\geq& 1/2 + \lbs h
\end{eqnarray*}

\section{Justifying Claims in the Active Upper Bounds}

\textbf{Phase 1 $(k=1)$.} In the first phase of the algorithm, it is possible
that $\sigma \preceq R_e/n \mbox{ but } \succeq R_e e^{-n}$ - in other words the noise may be small enough that passive learning cannot make out that we are in the errors-in-variables setting, and then the passive estimator will get a point error of $\frac{C_1R_e}{n/E}$ in each of those epochs (as if there is no feature noise).
This point error is to the best point in epoch $e$, which we can prove by
induction is the true threshold $t$ 
with high probability. Since it trivially holds in the first epoch ($t \in D_1 = [-1,1]$), we assume that it is true in epoch $e-1$. Then, in epoch
$e$, the true threshold $t$ is still the best point if the estimator $x_{e-1}$ of epoch $e-1$ was within $R_e$ of $t$, or in other words if
$|x_{e-1} - t|\leq R_{e}$. This would definitely hold if
$\frac{C_1R_{e-1}}{n/E} \leq R_e$
i.e.
$n \geq 2C_1E = 2C_1\lceil \log (1/\sigma) \rceil$, which is true since $\sigma \succ  \exp\{-n/2C_1\}$.
However, the algorithm
cannot stay in this phase of $\sigma \preceq R_e/n$ this until 
the last epoch since $\sigma > R_{E+1}=R_E/2$.

\textbf{Phase 2 ($k=1)$.} When $\sigma \succeq R_e/n$, WIDEHIST
gets an estimation error of $C_2\sqrt {\frac{R_e \sigma}{n/E}}$ in epoch $e$.
This error is the distance to the best point in epoch $e$, which is $t$ by the following similar induction. 
In epoch $e$, $t$ is still the best point only if $|x_{e-1} - t|\leq R_{e}$, i.e. 
$C_2^2 \frac{R_{e-1} \sigma}{n/E} \leq R_e^2$
i.e.
$nR_e ~\geq~ 2C_2^2 E\sigma$
which holds
since $R_e > \sigma$ for all $e \leq E$ and since $n \geq 2C_2^2 E$  ($\sigma \succ  \exp\{-n/2C_2^2\}$ implies $E \leq n/2C_2^2$).

The final error of the algorithm is
is $\sqrt {\frac{R_E \sigma}{n/E}} = \tilde{O}( \frac{\sigma}{\sqrt n} )$ since $R_E < 2\sigma$.

\paragraph{Explanation for $k>1$}
Assume $\sigma \succ n^{-\frac1{2k-2}}$, otherwise active learning won't notice the feature noise, and so $\log(1/\sigma) \leq \tfrac{\log n}{(2k-2)}$. Choose total epochs $E = \lceil\log (\tfrac{1}{\sigma}) \rceil \leq \frac{\log n}{(2k-2)} \leq C\log n$ for some $C$.  In each epoch of length $n/E$ in a region of radius $R_e=2^{-e+1}$, 
we get a passive bound of $C_1 \sqrt {\frac{R_e }{\sigma^{2k-3} n/E}}$ whenever  $\sigma > (\frac{R_e }{n})^{\frac1{2k-1}}$~. (This must happen at some $e \leq E= \lceil\log (\tfrac{1}{\sigma}) \rceil$ because  $R_E = 2^{-E+1} < 2\sigma < \sigma \sigma^{2k-2} n$ since $\sigma \succ n^{-\tfrac1{2k-2}}$ and hence in the last epoch 
$\sigma > (\frac{R_E }{n})^{\frac1{2k-1}}$.) By the same logic as for $k=1$, we need to verify that $|x_{e-1} - t| \leq R_e$ so that if $t$ was in the search space in epoch $e-1$ then it remains the in the search space in epoch $e$, i.e. we want to verify 
$C_1^2 \frac{R_{e-1}  }{ \sigma^{2k-3} n/E} \leq R_e^2 \Leftrightarrow \sigma^{2k-2} R_{e} \geq \frac{2C_1^2 E}{n} \sigma $
which is true since $R_{e} \geq \sigma$ and  $\sigma^{2k-2} > 2C_1^2 E / n$~. (By choice of $E= \lceil\log (\tfrac{1}{\sigma}) \rceil $, $R_e \geq R_E \geq \sigma \geq R_{E+1} $~. Since  $\sigma \succ n^{-\frac1{2k-2}}$ we get $\sigma^{2k-2} > 2C_1^2 E / n$ since $E \leq C\log n$~.)

The final point error is given by the passive algorithm in the last epoch as  $\sqrt {\frac{R_E }{\sigma^{2k-3} n/E}}$; since $R_E ~<~ 2\sigma$ and $E \leq C\log n$, this becomes $\preceq \frac1{\sigma^{k-2}} \sqrt {\frac1{ n}}$~.

\end{document}